\documentclass{article}

\PassOptionsToPackage{numbers, compress}{natbib}
 \usepackage[preprint]{neurips_2026}

\usepackage[utf8]{inputenc} 
\usepackage[T1]{fontenc}    
\usepackage{hyperref}       
\usepackage{url}            
\usepackage{booktabs}       
\usepackage{amsfonts}       
\usepackage{nicefrac}       
\usepackage{microtype}      
\usepackage{xcolor}         
\usepackage{graphicx}
\usepackage{amsmath}
\usepackage{multirow}
\usepackage[numbers]{natbib}
\usepackage{enumerate}
\usepackage{enumitem}

\usepackage{etoc}
\usepackage{titletoc}

\usepackage{amssymb}
\usepackage{amsthm}
\usepackage[table]{xcolor}
\usepackage{caption}
\usepackage{mathtools}
\usepackage{subcaption}
\usepackage{placeins}
\usepackage{algorithm}
\usepackage{algorithmicx}
\usepackage[noend]{algpseudocode}
\usepackage{subcaption}

\title{Commanding Humanoid by Free-form Language:\\
A Large Language Action Model with \\
Unified Motion Vocabulary}

\author{%
  Zhirui Liu$^{1,2,*}$ \hspace{1em} Kaiyang Ji$^{1,2,}$\thanks{Equal contribution. $\dag$Corresponding author.} \hspace{1em} Ke Yang$^{1,2}$ \hspace{1em}  Yahao Fan$^{1,2}$ \hspace{1em} Jingyi Yu$^{1}$ \\ \hspace{1em} \textbf{Ye Shi$^{1,2}$ \hspace{1em} Jingya Wang$^{1,2,\dag}$} \\ 
  \vspace{1pt}\\
  $^1$ShanghaiTech University \hspace{1em}
  $^2$InstAdapt\\
  \vspace{1pt}\\
  \texttt{ \{liuzhr2025, jiky2024, fanyh12024\}@shanghaitech.edu.cn} \\
  \texttt{ky2276@outlook.com} \\
  \texttt{\{yujingyi, shiye, wangjingya\}@shanghaitech.edu.cn} \\
}

\begin{document}

\maketitle

\begin{abstract}
  Enabling humanoid robots to follow free-form natural language commands is a critical step toward seamless human-robot interaction and general-purpose embodied AI. However, existing methods remain limited, often constrained to simple instructions or forced to sacrifice motion diversity for physical plausibility. To address this gap, we present Humanoid-LLA, a Large Language Action model that translates unconstrained natural language directly into executable whole-body motions for humanoid robots. Our approach tackles two core challenges: paired language-humanoid motion data scarcity and physical instability. First, we bridge high-level language semantics with physically-grounded control by learning a unified human-humanoid motion vocabulary. Second, we introduce a novel two-stage fine-tuning framework that begins with supervised motion Chain-of-Thought learning, followed by reinforcement learning refined with physical feedback to ensure robustness and stability. Extensive evaluation in simulation and real-world cross-embodiment experiments demonstrates that Humanoid-LLA achieves superior generalization to novel language commands and diverse motion generation while maintaining high physical fidelity. Project page: \url{https://humanoidlla.github.io/}.

\end{abstract}

\section{Introduction}
\label{sec:intro}

Enabling humanoid robots to follow free-form language instructions while maintaining physically stable whole-body behaviors remains a critical challenge for embodied intelligence. Recent progress in large language models (LLMs) \citep{shao2024deepseekmath, yang2025qwen3} and vision-language-action models (VLA) \citep{kimopenvla, bjorck2025gr00t, xu2024humanvla, xue2025leverb,ding2025humanoid} has begun to bridge foundation models with embodied systems. Despite VLAs having shown strong results in manipulation tasks, free-form language control for humanoid robots remains challenging due to the high
degree of freedom and complex dynamics inherent in humanoid robots, specifically: (i) paired data of language instructions and physically executable whole-body motions for humanoids is expensive and difficult to collect at scale, and prior pipelines further discard a large fraction of human text–motion data due to non-executability on a specific humanoid; (ii) whole-body humanoid motions are highly sensitive to physical instability, where small errors can be amplified through contact dynamics and lead to falls.

Recent pioneering works have begun to explore simple language-conditioned whole-body control. These efforts generally fall into two categories. The first and more established paradigm relies on motion mimicking frameworks \cite{he2025omnih2o, yue2025rl, xu2024lagoon}. In this pipeline, human motions are first generated from text using off-the-shelf models \citep{tevethuman, zhang2023generating, jiang2024motiongpt, hymotion2025} and subsequently retargeted to humanoid robots. However, directly coupling generation with retargeting \cite{joao2025gmr,yang2025omniretarget} introduces significant challenges. A key issue is that text-to-human generators, trained solely on human motion distributions, often produce outputs that violate humanoid dynamics, necessitating labor-intensive curation for physical feasibility.
The second paradigm utilizes distillation frameworks\citep{shao2025langwbc, xue2025leverb}, which transfer knowledge from a privileged tracking controller to a text-conditioned policy. While this method achieves strong physical fidelity, it typically compresses both semantic understanding and motor control into a single variational model. This compression often weakens the model's grounding in language and blurs precise action selection.
Critically, both paradigms are fundamentally limited to mapping simple instructions to motions. They lack the deep comprehension, reasoning, and generative capabilities of modern large language models, failing to translate complex, high-level linguistic commands into nuanced and contextually appropriate whole-body behaviors.

To address these limitations, we introduce Humanoid-LLA, a Large Language Action model designed for free-form text-to-humanoid whole-body control. Our model directly maps expressive natural language instructions into executable humanoid actions, leveraging its strong reasoning and analytical capabilities to interpret and decompose high-level commands.

To overcome the fundamental challenge of data scarcity in humanoid whole-body control, we propose building a unified human–humanoid discrete motion vocabulary that maximizes the utility of existing motion data. Importantly, this “data scarcity” is not about the lack of human text–motion data, but about the limited amount of paired trajectories that are physically executable on a specific humanoid: prior pipelines typically filter out a substantial fraction of human motions as non-executable or rely on costly curation, shrinking motion–language coverage and weakening broad language grounding. Our unified vocabulary enables leveraging the full human corpus to learn language–motion alignment, while using the humanoid-executable subset to learn physical grounding. Specifically, we construct this shared vocabulary through joint quantization of paired human motions and their retargeted humanoid counterparts. This process enforces bidirectional cross-embodiment reconstruction, ensuring that the same discrete token consistently represents the same motion primitive across both human and humanoid embodiments.

Moreover, to enhance physical stability for robust execution, we introduce two key components: (1) A vocabulary-directed controller, distilled from a privileged tracking teacher policy, serves as the critical bridge that converts an LLM-generated vocabulary sequence into stable motor execution. This controller provides implicit vocabulary tracking, eliminating the need for online retargeting at inference time. (2) Reinforcement learning fine-tuning guided by humanoid physical feedback. This process involves executing proposed vocabulary sequences in simulation and using a physical reward signal to steer the model’s output toward the humanoid robot’s feasible motion manifold. Extensive evaluations in both simulation and real-world environments
demonstrate compelling language generalization capabilities while maintaining high physical fidelity. 

We summarize our core contributions as follows:

\begin{itemize}[leftmargin=2em, rightmargin=2em]
    \item We present Humanoid-LLA, a Large Language Action Model that enables the first free-form text-to-humanoid whole-body control, mapping expressive natural language directly to executable humanoid actions. 
    

    \item We propose a unified human–humanoid motion vocabulary learned via cross-embodied tokenization, providing a shared learnable latent space for humanoid control, LLM modeling, and cross-embodiment transfer, thereby mitigating embodiment mismatch and data scarcity in whole-body humanoid control.
    
    \item To bridge the gap between high-level instruction understanding and physically-constrained execution, we propose a systematic two-stage fine-tuning framework. By sequentially combining supervised motion CoT learning and RL with physical feedback, Humanoid-LLA acquires a unified capability for semantic reasoning and physical stability, enabling robust generalization under real-world dynamics.
    
\end{itemize}

\begin{figure}[t] 
    \centering
    \includegraphics[width=\linewidth]{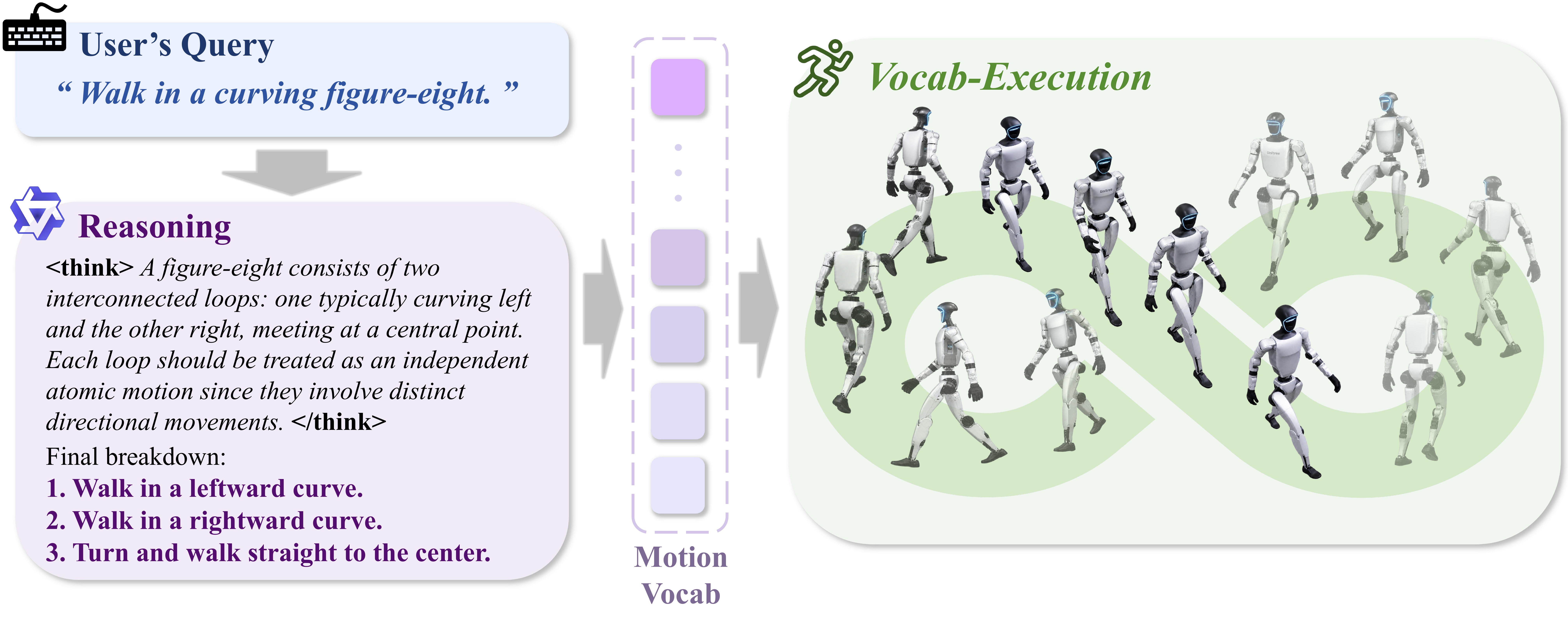}
    \caption{An illustration of Humanoid-LLA. Given a high-level and abstract command, Humanoid-LLA first uses natural language to decompose the task into several sub-tasks, and then a downstream controller is responsible for sequential execution, yielding semantically aligned and physically faithful whole-body behaviors.}
    \label{fig:teaser}
\end{figure}

\section{Related Work}
\label{sec:related work}

\paragraph{Text-to-Motion Generation.} Text-to-motion work typically treats motion synthesis as conditional sequence modeling, producing long-horizon pose trajectories from language. Diffusion-based methods \citep{tevethuman,chen2023executing,zhang2024motiondiffuse,karunratanakul2023guided} yield diverse and high-quality motions but remain expensive and difficult to steer, while GPT-based generators \citep{zhang2023generating,jiang2024motiongpt,ouyang2025motion} improve efficiency and temporal consistency yet are prone to discretization and representation artifacts. Recent efforts \citep{yuan2023physdiff, serifi2024robot, han2025reindiffuse} inject physical priors by projecting kinematic outputs into feasible states or by training with reward surrogates and RL controllers, but the resulting motions are still often generated in a human-centric space and may not be directly executable on a humanoid. In contrast, we aim to learn language–motion grounding in a unified human–humanoid discrete vocabulary, so that scaling with abundant human data does not decouple generation from downstream humanoid feasibility.

\paragraph{Physics-Based Control with Language Interface.}
To correct kinematic artifacts such as foot sliding and implausible contacts, physics-based animation trains controllers in simulation to enforce dynamical consistency, enabling robust imitation and skill composition \citep{peng2018deepmimic, peng2021amp, peng2022ase, tessler2024maskedmimic}. However, specifying rich behaviors purely via low-level rewards or task-specific controllers can be cumbersome, motivating language-conditioned control of physically simulated agents \cite{juravsky2022padl, juravsky2024superpadl}. Recent systems combine linguistic interfaces with motion codebooks, physics-aware imitation, or closed-loop plan-and-imitate schemes \cite{yao2024moconvq,truong2024pdp,tevetclosd,wu2025uniphys}, pointing toward closed-loop frameworks that couple semantic intent with physical execution. This work is inspired by this direction but differs by introducing a unified human–humanoid vocabulary interface and a vocab-directed controller, enabling token-level execution as a stable bridge from language generation to real-world physical execution.

\paragraph{Language-Conditioned Humanoid Whole-Body Control.}
Progress in humanoid control has improved retargeting, teleoperation, and sim-to-real robustness \cite{Luo2023PerpetualHC, joao2025gmr, yang2025omniretarget, he2025omnih2o, ze2025twist, ze2025twist2, ben2025homie, li2025amo, li2025clone, he2024learning, cheng2024expressive, ji2024exbody2, he2025hover, he2025asap, liao2025beyondmimic, yin2025unitracker, luo2025sonic}, yet language-conditioned whole-body control \cite{shi2025almi, shao2025langwbc, yue2025rl, li2025language, wang2025sentinel, jiang2025uniact, li2026w1, luo2025sonic, xie2026textop, bao2026phygile, yuan2026roboforge} still faces a semantic-to-physical gap. Early explorations such as Harmon \cite{jiang2025harmon}, UH-1 \cite{11203143}, and ALMI \cite{shi2025almi} improve text--humanoid motion corpora but struggle with real-world deployment. LangWBC \cite{shao2025langwbc} boosts physical fidelity at the cost of language generalization, and RLPF \citep{yue2025rl} finetunes an LLM-based generator with physical feedback, but operates in human-motion space and can restrict diversity due to conservative binary reward.

More recently, RoboGhost \cite{li2025language} adopts retargeting-free latent-driven control, and SENTINEL \cite{wang2025sentinel} trains an end-to-end language--action model on a large tracking-built dataset. UniAct \cite{jiang2025uniact} uses shared discrete quantization for multi-modal motion generation. 
However, these methods largely decouple generation from execution, offering limited dynamic feedback to the generator, and underexplore complex commands requiring semantic decomposition. In contrast, we use simulation-based rewards to realize closed-loop physics-aware token generation.
Additionally, our unified motion vocabulary enables human-to-cross-embodiment transfer, with extensive real-world experiments on Unitree G1 and Booster T1 demonstrating the paradigm's effectiveness and generalizability.

\section{Method}
\begin{figure*}[t] 
    \vskip 0.2in
    \centering
    \includegraphics[width=\linewidth]{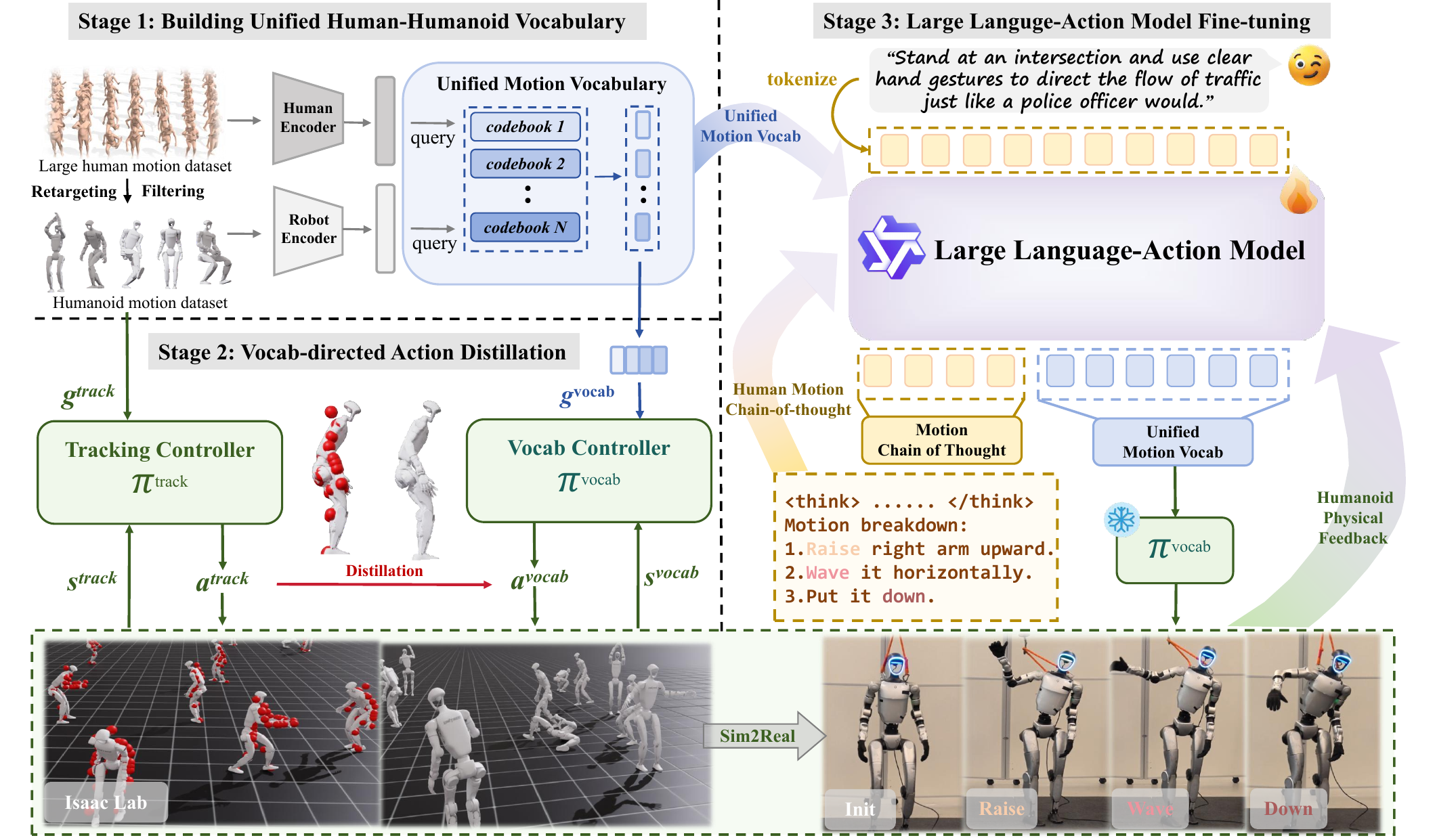}
    \caption{An overview of Humanoid-LLA. In stage one, we build a unified motion vocabulary leveraging a large-scale paired human and humanoid motion dataset. With a kinematic humanoid motion goal and its corresponding vocab retrieval, we distill a vocab-directed humanoid student controller from a teacher tracking controller. The first two stages enable stage three to acquire various humanoid feedback directly from physical simulation without decoding, making our LLA enhanced with high physical fidelity and language generalization. }
    \label{fig:prism}
\end{figure*}

Our framework, as shown in Fig.~\ref{fig:prism}, consists of three tightly connected components: building unified human–humanoid motion vocabulary (Sec.~\ref{sec:quantization}), distilling vocabulary-directed policy (Sec.~\ref{sec:distillation}), and fine-tuning large language-action model (Sec.~\ref{sec:LLA}). The first two components serve as essential prerequisites that make the integrated reasoning in the third component possible. Next, we introduce each component, highlighting its role within the overall framework. 

\subsection{Unified Human-Humanoid Vocabulary}
\label{sec:quantization}
\paragraph{Humanoid Motion Canonicalization.}
\label{sec:canonicalization}
To enable cross-embodiment alignment, we canonicalize humanoid motions into a structured, root-centered representation that is compatible with the commonly used human motion parameterization \cite{guo2022generating}. The representation aggregates joint configuration together with essential kinematic signals, and is normalized to a consistent coordinate frame for learning a shared vocabulary. 

\paragraph{Implicit Partitioning Tokenization.}
We aim to learn a unified tokenizer that maps human and retargeted humanoid motions into the same discrete vocabulary, ensuring that identical tokens carry consistent semantics across embodiments. 
This vocabulary serves as a compact interface for language-model reasoning and enables scalable training with abundant human motion data.
  
A discrete interface, however, may overcompress fine-grained kinematics required by stable whole-body control. 
To mitigate this information bottleneck and avoid manual body-part-based partitioning of motion representations, we adopt VQ-VAE \cite{van2017neural} with implicit partitioning \cite{unitok}, where each latent vector is factorized into sub-blocks and quantized by multiple sub-codebooks. This factorization increases representational bandwidth without requiring a single giant codebook, thereby preserving motion details that are critical for stable execution.

\paragraph{Cross-embodiment Optimization.}
We optimize the dual-branch VQ-VAE by combining intra-modal and cross-modal reconstruction objectives. A sequence of human motion $\mathbf{m} \in \mathbb{R}^{T \times d_h}$ and humanoid motion $\mathbf{m}
 \in \mathbb{R}^{T \times d_r}$ are first encoded into latent features $\mathbf{z
 }^{\text{h}}=\mathcal{E}_{\text{human}}(\mathbf{m}^{\text{h}})$ and $\mathbf{z}^{\text{r}}=\mathcal{E}_{\text{robot}}(\mathbf{m}^{\text{r}})$, which are partitioned into sub-blocks and quantized by multiple codebooks to yield discrete tokens $\hat{\mathbf{z}}^{\text{h}}$ and $\hat{\mathbf{z}}^{\text{r}}$. These tokens are then decoded back to the motion space by modality-specific decoders $\mathcal{D}_{\text{human}}$ and $\mathcal{D}_{\text{robot}}$, producing both self-reconstructions ($\hat{\mathbf{m}}^{\text{h}}$, $\hat{\mathbf{m}}^{\text{r}}$) and cross-reconstructions ($\hat{\mathbf{m}}^{\text{r}\leftarrow\text{h}}$, $\hat{\mathbf{m}}^{\text{h}\leftarrow\text{r}}$). Beyond standard self-reconstruction within each modality \citep{zhao2025smap}, we additionally enforce cross-modal reconstruction, such that a token obtained from either modality is decoded into the same motion primitive, which is crucial for achieving semantically unified tokenization.
The training objective is defined as:
\begin{equation}
\label{eq:1}
\mathcal{L} = \mathcal{L}_{\text{intra}} + \alpha \mathcal{L}_{\text{commit}} + \beta \mathcal{L}_{\text{cross}} ,
\end{equation}
where $\mathcal{L}_{\text{intra}}$ is the intra-modal reconstruction loss, $\mathcal{L}_{\text{cross}}$ penalizes discrepancies in cross-modal reconstruction, and $\mathcal{L}_{\text{commit}}$ is the commitment loss. Balancing coefficients $\alpha$ and $\beta$ control the trade-off between fidelity and codebook consistency. See Appendix for more architectural and training details.

\subsection{Vocabulary-directed Humanoid Action Distillation}
\label{sec:distillation}
With unified motion vocabulary in Sec.~\ref{sec:quantization}, we next bridge the gap between kinematic motion primitives and physical control through a vocabulary-directed distillation process. Following the teacher–student paradigm used in recent whole-body controllers\citep{he2025omnih2o, yin2025unitracker, tessler2024maskedmimic, yin2025visualmimic}, we train a privileged teacher policy to track continuous humanoid-retargeted motions with high fidelity and then distill its behavior into a vocabulary-directed student policy that relies on motion tokens. This stage shifts the control input from dense reference trajectories to the compact motion language of tokens, enabling the humanoid to execute token sequences output by the language model in Sec.~\ref{sec:LLA}.

\paragraph{Fully-constrained Teacher Controller.}
\label{sec:teacher}
We follow the goal-conditioned reinforcement learning framework to train a fully-constrained teacher tracking policy $\pi^{\text{track}}$ that tracks dense humanoid-retargeted reference states. At timestep $t$, the controller observes humanoid proprioception $\mathbf{s}_t$ and a goal state $\mathbf{g}_t^{\text{track}}$ comprising kinematic reference motion, and computes target joint positions $\mathbf{a}_t$ for the PD controller.

The teacher proprioception $\mathbf{s}_t$ consists of the current root linear velocity $\mathbf{\dot{p}}_t^{\text{root}} \in \mathbb{R}^3$, root angular velocity $\omega_t^{\text{root}} \in \mathbb{R}^3$, joint positions $\mathbf{q}_t \in \mathbb{R}^{n_j}$, joint velocities $\mathbf{\dot{q}_t} \in \mathbb{R}^{n_j}$ and the previous action history $\mathbf{a}_{t-1} \in \mathbb{R}^{n_j}$ with respect to the robot's local coordinate frame:
\begin{equation}
    \mathbf{s}_t =
    \Big[
    \mathbf{\dot{p}}_t^{\text{root}}, \  \omega_t^{\text{root}}, \ \mathbf{q}_t, \ \mathbf{\dot{q}}_t, \ \mathbf{a}_{t-1}
    \Big].
\end{equation}
And for tracking goal observation $\mathbf{g}^{\text{track}}_t$, we track relative body pose instead of absolute poses following a previous tracking framework \citep{liao2025beyondmimic}:
\begin{equation}
    \mathbf{g}^{\text{track}}_t =
    \Big[
        \hat{\mathbf{q}}_{t+1}, \ \hat{\dot{\mathbf{q}}}_{t+1}, \ \hat{\mathbf{p}}_{t+1}^{\text{root}} - \mathbf{p}^{\text{root}}_{t}, \ \hat{\theta}^{\text{root}}_{t+1} \ominus \theta_t^{\text{root}}
    \Big],
\end{equation}
where $\ominus$ denotes the difference between two rotations. The policy action $\mathbf{a}_t$ is the normalized robot target joint positions, which are residual targets for nominal joint configuration.

For policy training, Proximal Policy Optimization (PPO)~\citep{schulman2017proximal} algorithm is used to maximize the accumulated reward $r_t = \mathbb{E}[\sum_{t=1}^{T}\gamma^{t-1}r_t]$. We design the reward $r_t$ as a weighted sum of task rewards, regularization, and penalty. Details can be found in the Appendix.

\paragraph{Vocabulary-directed Student Controller.}
\label{sec:student}
After the fully-constrained teacher controller is trained, we distill $\pi^{\text{track}}$ into a vocabulary-directed student policy. Let the unified tokenizer (Sec.~\ref{sec:quantization}) provide a motion vocab window $\hat{\mathbf{z}}_{1:T}^{\text{vocab}}$, we aim to train a student policy $\pi^{\text{vocab}}$ that can generate full body actions satisfying these given motion vocabulary commands. To solve this ambiguity, we follow \cite{tessler2024maskedmimic, tessler2025maskedmanipulator} and model $\pi^{\text{vocab}}$ as a Conditional Vatiational Autoencoder (CVAE)~\citep{kingma2013auto} consisting of a vocabulary prior $\rho$, a residual encoder $\mathcal{E}$ and an action decoder $\mathcal{D}$. At timestep $t$, the motion vocab observation of the student controller is:
\begin{equation}
    \mathbf{g}_t^{\text{vocab}} = \Big[
    \mathcal{M}(\mathbf{g}_t^{\text{track}}), \ \hat{\mathbf{z}}^{\text{vocab}}_t 
    \Big],
\end{equation}
where $\mathcal{M}(\cdot)$ is a random masking function and $\hat{\mathbf{z}}^{\text{vocab}}_t$ is the current motion vocabulary in Sec.~\ref{sec:quantization}. The vocabulary prior is modeled as a Gaussian distribution over latents given the observed vocab constraints:
\begin{equation}
    \rho(z_t | \mathbf{s}_t, \mathbf{g}_t^{\text{vocab}} ) = \mathcal{N}\left(\mu^{\rho}(\mathbf{s}_t, \mathbf{g}_t^{\text{vocab}}), \ \sigma^{\rho}(\mathbf{s}_t, \mathbf{g}_t^{\text{vocab}})\right).
\end{equation}
The encoder $\mathcal{E}$ is modeled as a residual to the prior that outputs a latent distribution given the full-constraint teacher observation $\mathbf{g}_t^{\text{track}}$~\citep{yao2022controlvae}:
\begin{align}
\mathcal{E}(z_t \mid \mathbf{s}_t, \mathbf{g}_t^{\text{track}})
= 
\mathcal{N}\Big(
    \mu^{\rho}(\mathbf{s}_t, \mathbf{g}_t^{\text{vocab}})
    + \mu^{\mathcal{E}}(\mathbf{s}_t, \mathbf{g}_t^{\text{track}}), \ \sigma^{\mathcal{E}}(\mathbf{s}_t, \mathbf{g}_t^{\text{track}})
    \Big).
\end{align}

Based on the Dataset Aggregation (DAgger) algorithm~\citep{ross2011reduction}, we train $\pi^{\text{vocab}}$ from $\pi^{\text{track}}$ with motion vocab labels within the same motion dataset. The training objective is to minimize the difference between reference action and student action as well as the KL divergence between encoder distribution $p_{\mathcal{E}}$ and prior distribution $q_{\mathcal{\rho}}$:
\begin{align}
\mathcal{L}_{\pi^{\text{vocab}}}
&= \| a^{\text{track}}_{t} - a_t^{\text{vocab}} \|_2^2 + \lambda_{\text{KL}}
\Big(
    p_{\mathcal{E}}(z_t \mid \mathbf{s}_t, \ \mathbf{g}_t^{\text{track}})
    \ \big\|\
    q_{\rho}(z_t \mid \mathbf{s}_t, \ \mathbf{g}_t^{\text{vocab}})
\Big),
\end{align}
where $a^{\text{track}}_{t}$ is the reference action from $\pi^{\text{track}}$, $a_t^{\text{vocab}}$ is the student action sampled from $\mathcal{D}(a_t^{\text{vocab}}|\mathbf{s}_t, \mathbf{g}_t^{\text{vocab}})$ and $\lambda_{\text{KL}}$ is the hyperparameter for balancing reconstruction and regularization. 

\subsection{Large Language-Action Model}
\label{sec:LLA}
In this section we show how, building upon Sec .~\ref {sec:quantization} and Sec.~\ref {sec:distillation}, our framework realizes a direct mapping from highly abstract language descriptions to physically executable robot actions without decoding and retargeting during inference. Sec.~\ref {sec:distillation} serves as the key intermediate: a low-level controller distilled to follow latent motion tokens, seamlessly linking latent motion token generation and physics-based action execution. The remainder of this section details the training procedure.

\paragraph{Supervised Fine-tuning with Augmented Human Data.}
To solve the limited expressivity of textual annotations in existing datasets \cite{guo2022generating, punnakkal2021babel, hwangsnapmogen}, prior approaches \cite{ouyang2025motion} have resorted to LLM \cite{shao2024deepseekmath} to decompose abstract motion descriptions. However, such methods often cause motion-language misalignment since a single caption may correspond to multiple plausible motions. Inspired by recent works \cite{cui2024anyskill} that demonstrate the effectiveness of visual signal for motion-text understanding, we employ a vision–language model \cite{Qwen2.5-VL} with rendered motion sequences as motion context, enabling it to produce more accurate motion chain-of-thought (CoT). We use motion CoT only for abstract or compositional instructions, and atomic commands are passed through unchanged.

Given the augmented large-scale human motion–text datasets,
we formulate motion token generation as an autoregressive, text-conditioned language modeling task, where a motion sequence is represented as a series of discrete tokens from the unified codebook $\mathcal{Z}=\{\langle cb_{i,j}\rangle\}$, with $i$ indexing the sub-codebook and $j$ the token entry. The input is the textual description $\mathbf{w}$, and the supervision target $\mathbf{y}=(y_1,\dots,y_L)$ is constructed by concatenating motion chain-of-thought with the ground-truth motion tokens from the pretrained tokenizer. The model is trained with the standard next-token prediction loss:
\begin{equation}
    \mathcal{L}_{\text{SFT}}
    = -\mathbb{E}_{(\mathbf{w},\mathbf{y})\sim\mathcal{D}}
    \;\sum_{t=1}^{L}
    \log P_{\phi}\big(y_t \mid \mathbf{w}, y_{<t}\big),
\end{equation}
where $\phi$ are the model parameters. Importantly, we perform SFT for language–motion semantic alignment using the full human text–motion corpus under the unified vocabulary; physical executability is not required at this stage. We only enforce humanoid feasibility in the later stage through our pretrained vocab-controller and physics-based RL fine-tuning on the G1-executable subset.

\paragraph{RL Fine-tuning with Humanoid Feedback.}
Large models are commonly adapted to downstream tasks with reinforcement learning, resulting in policies that better match task-specific requirements. We adopt Group Relative Policy Optimization (GRPO) \citep{shao2024deepseekmath}, a variant of PPO \citep{schulman2017proximal} that avoids training a separate critic by sampling a group of candidate outputs $y^{(1:K)}$ for each input prompt $x$, assigning each a scalar reward, and normalizing rewards within the group to obtain relative advantages. This encourages the policy to prefer better-than-average candidates without requiring an explicit value function. The policy is optimized with a clipped surrogate objective regularized toward a reference model:
\begin{align}
\mathcal{L}_{\text{GRPO}}(\phi)
&= 
-\,\mathbb{E}_{x}\,
\mathbb{E}_{y^{(1:K)} \sim \pi_{\phi}}
\Bigg[
    \frac{1}{K}\sum_{k=1}^{K}
    \min\!\Big(
        r_{k}\,\tilde{A}_{k}, \ \mathrm{clip}(r_{k};\,1-\epsilon,\,1+\epsilon)\,\tilde{A}_{k}
    \Big)
\Bigg]
+\,\beta_{\text{KL}}\,\mathcal{L}_{\text{KL}},
\end{align}

where $x$ is the input prompt, $y^{(1:K)}$ are $K$ sampled candidate sequences, $r_k$ is the likelihood ratio between the current and reference policies, and $\tilde A_k$ is the group-normalized advantage. The KL term $\mathcal{L}_{\text{KL}}$ constrains the policy to stay close to a reference model. This formulation provides a stable and efficient way to fine-tune with humanoid feedback, injecting physical priors into token generation.

Unlike prior work that emphasizes kinematic fidelity \citep{ouyang2025motion, yue2025rl}, we stress the importance of dynamics-level consistency for real-world deployment.  RLPF \citep{yue2025rl} employs a binary simulator-tracking reward, which ensures executability but often reduces motion diversity, as the policy tends to favor conservative behaviors that are easy to track.  To address this, we design a reward scheme that combines high-level distributional objectives with low-level simulator-based tracking signals, achieving motions that are both physically robust and expressively varied.

\paragraph{Physical Fidelity Reward Design.}
The overall reward is a weighted sum of a binary format reward and a continuous physical fidelity reward. The format reward acts as a prerequisite: the model must first learn \textit{how to answer} (i.e., producing valid structured outputs) before it can effectively learn \textit{how to answer well} (i.e., generating physically and semantically aligned motions). Concretely, the format reward checks two requirements: (i) the response must follow a structured template beginning with \texttt{<think>...</think>} and followed by \texttt{<motion>...</motion>}; and (ii) within the motion segment, motion tokens must appear in cyclic sub-codebook order (\texttt{cb0}→\texttt{cb1}→…→\texttt{cb(N-1)} repeatedly). We define it as
\begin{equation}
    r_{\text{format}} = \mathbb{I}\{\text{requirements satisfied}\}.
\end{equation}
The physical fidelity reward is composed of a distributional term and a tracking term. The distributional reward encourages motion distribution generated by the vocab-controller to match the distribution of physically plausible trajectories and to align semantically with the paired motion descriptions. Using contrastive motion encoder $\phi_{\text{m}}(\cdot)$ and text encoder $\phi_{\text{t}}(\cdot)$ \citep{guo2022generating} trained on physically plausible humanoid datasets, we define distributional reward as:
\begin{equation}
\label{eq:11}
r_{\text{dist}}
=
\exp\!\big(
    -\lambda_{m}\,
    \|\phi_{\text{m}}(\mathbf{m}_{\text{gen}})
    -\phi_{\text{m}}(\mathbf{m}_{\text{ref}})\|_2
\big)
+\,\exp\!\big(
    -\lambda_{t}\,
    \|\phi_{\text{m}}(\mathbf{m}_{\text{gen}})
    -\phi_{\text{t}}(\mathbf{w}_{\text{ref}})\|_2
\big),
\end{equation}
where the two terms collectively measure motion fidelity and semantic fidelity. $\mathbf{m}_{\text{gen}}$, $\mathbf{m}_{\text{ref}}$ and $\mathbf{w}_{\text{ref}}$ represent the motion rolled out in simulation by the vocab-controller, ground-truth motion and paired motion description, respectively. $\lambda_{m}$, $\lambda_{t}$ are balancing coefficients. 

The tracking reward measures how well a generated token sequence can be executed in simulation by the distilled vocab-controller (Sec.~\ref{sec:distillation}). We evaluate the simulated rollout with a position reward term $\text{r}_{\text{pos}}$ and an acceleration reward term $\text{r}_{\text{acc}}$:
\begin{equation}
    \label{eq:12}
    r_{\text{track}}
    =
    r_{\text{pos}}
    +
    r_{\text{acc}}.
\end{equation}

Finally, the overall reward is calculated as $r = r_{\text{format}} + r_{\text{dist}} + r_{\text{track}}$.
For more details about the Large Language-Action Model design, see Appendix.

\section{Experiments}
\label{exp}
\subsection{Experiment Setup}
\paragraph{Dataset.} 
We conduct extensive experiments on the text-annotated subset of AMASS~\citep{AMASS:ICCV:2019,guo2022generating}, consisting of $\sim$27K motion sequences. For each motion sequence, we employ \texttt{mink} \citep{zakka10mink} to retarget human motions into corresponding kinematic humanoid motions. We then use our privileged tracking teacher to remove physically infeasible motions, yielding a G1-executable subset that retains approximately 70\% of the full dataset. For each raw motion description, we generate a motion chain-of-thought using Qwen2.5-VL \cite{Qwen2.5-VL}, yielding a paired dataset with rich and coherent reasoning about the underlying motion, compared with the original short and generic descriptions. The choice of this dataset is motivated by two factors. First, AMASS is a mocap-based dataset, ensuring low noise compared with other datasets \cite{lin2023motion,fan2025go} curated from Internet data, thus enabling the model to better learn the latent alignment between motion and language. Second, text-annotated AMASS has been widely adopted in both human motion generation and humanoid whole-body control, which ensures standardized and fair comparison across methods.

\paragraph{Baselines.}
To comprehensively demonstrate the advantages of our model in terms of both motion quality and physical executability for text-to-humanoid, we compare against several state-of-the-art baselines: 1) \textbf{MDM+Retarget} \citep{tevethuman} kinematically retargets MDM-generated motion to humanoid robots. 2) \textbf{OmniH2O} \citep{tevethuman, he2025omnih2o} uses MDM to produce kinematic human motions followed by retargeting and an imitation policy. 3) \textbf{UH-1} \citep{11203143} trains a decoder-only transformer to map text descriptions into humanoid motion with a retargeted humanoid motion-text dataset. 4) \textbf{LangWBC} \citep{shao2025langwbc} distills a CVAE-based policy to simultaneously capture text semantics and sample actions.
 5) \textbf{RLPF} \citep{yue2025rl} explores physical feedback to constrain the kinematic LLM-based human motion generator, which is also followed by a post-process of motion retargeting and tracking. 
Besides text-to-humanoid, refer to Appendix for more experiments and ablation results for building unified motion vocabulary~\ref{sec:quantization} and distilling vocab-directed controller~\ref{sec:distillation}.
 6) \textbf{UniAct}\citep{jiang2025uniact} provides multi-modal unified motion generation, and we use its text-to-motion (T2M) branch as our baseline.

\paragraph{Evaluation Metrics.}
Most prior work on text-to-humanoid motion generation \citep{11203143, shao2025langwbc, shi2025almi, yue2025rl} reports either low-level physics tracking metrics or human-motion generation metrics, leaving no unified protocol directly defined on humanoid robots. To fill this gap, we design an evaluation that combines physics-based tracking measures with distributional generation metrics computed in humanoid motion space.
These two perspectives jointly capture executability, distributional fidelity, motion--language alignment, and diversity, thus discouraging models from producing only simple, easily executable motions at the expense of expressiveness.  
For the generation side, we report FID to measure distributional similarity against a physical humanoid motion set obtained by a goal-conditioned tracking policy (i.e., teacher controller in Sec.~\ref{sec:teacher}), MM-Dist and R-Precision to assess motion-language alignment, and Diversity (Div.) to evaluate variability. For the physics side, we measure success rate (Succ.), mean per-joint position error MPJPE (mm), velocity error $\text{E}_\text{vel}$ (mm/frame), and acceleration error $\text{E}_\text{acc}$ (mm/frame$^2$). 
Refer to Appendix for details.

\begin{figure*}[t] 
    \centering
    \includegraphics[width=\linewidth]{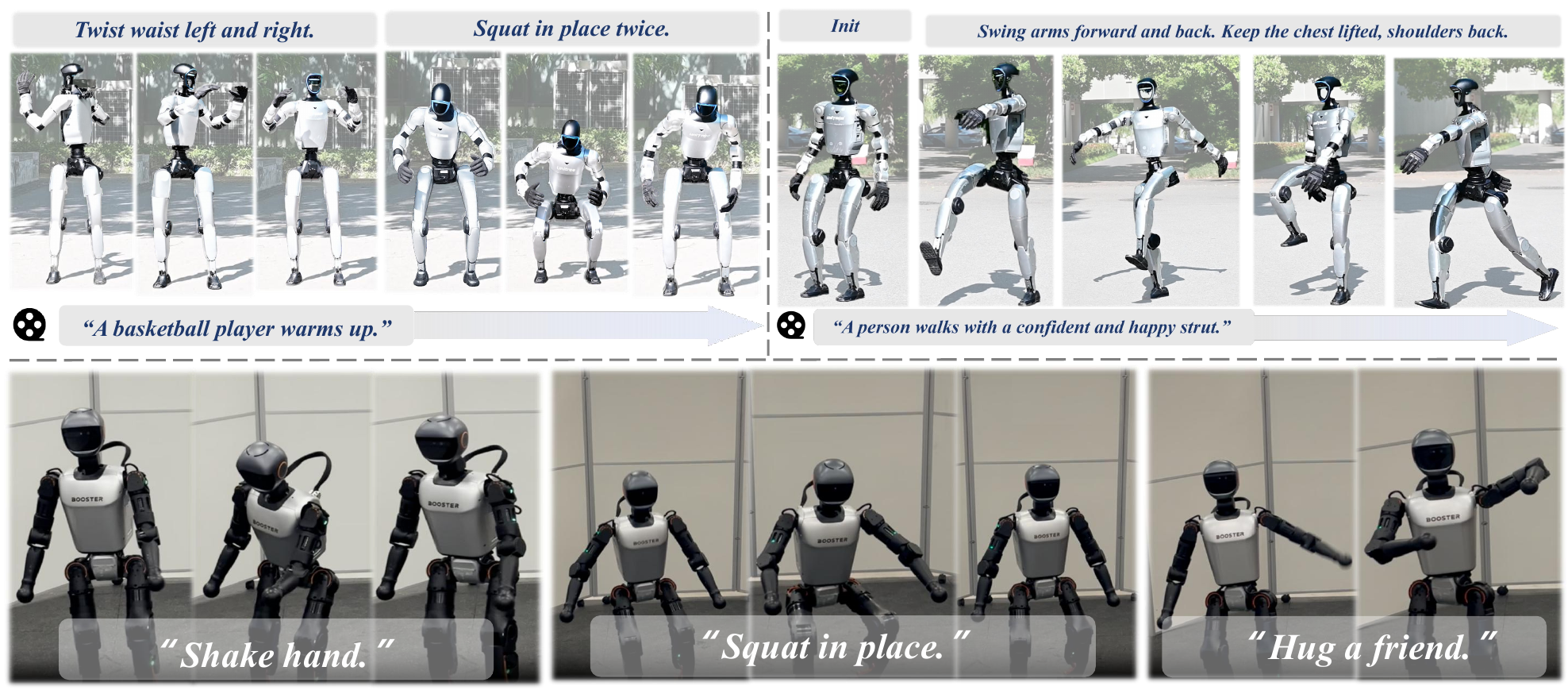}
    \caption{Real-world experiments on Unitree G1 (top) and Booster T1 (bottom). We test both simple instructions and abstract instructions that require decomposition. The results demonstrate that our method simultaneously achieves semantic consistency and physical robustness.}
    \vspace{-0.08in}
    \label{fig:real_result}
\end{figure*}

\begin{table}[t]
\caption{Text-to-humanoid results: generation metrics on the left and physics-based metrics on the right. R-Precision is reported at top-3. Arrows indicate the preferred direction.}
\centering
\scriptsize
\setlength{\tabcolsep}{3pt}
\renewcommand{\arraystretch}{0.95}

\begin{subtable}[t]{0.5\linewidth}
\centering
\begin{tabular}{l|cccc}
\toprule
Methods & FID$\downarrow$ & R-Prec.$\uparrow$ & MM-Dist$\downarrow$ & Div.$\rightarrow$ \\
\midrule
MDM+Retarget~\citep{tevethuman} & $11.759$ & $0.262$ & $6.599$ & $6.419$ \\
OmniH2O~\citep{he2025omnih2o} & $17.159$ & $0.222$ & $8.021$ & $5.868$ \\
UH-1~\citep{11203143} & $8.682$ & $0.295$ & $5.896$ & $6.749$ \\
LangWBC$^*$~\citep{shao2025langwbc} & $6.171$ & $0.320$ & $5.587$ & $6.031$ \\
UniAct-T2M~\citep{jiang2025uniact} & $8.513$ & $0.284$ & $6.031$ & $6.514$ \\
\midrule
\rowcolor[HTML]{EFEFEF}
Humanoid-LLA & $\mathbf{2.626}$ & $\mathbf{0.447}$ & $\mathbf{4.911}$ & $\mathbf{7.122}$ \\
\bottomrule
\end{tabular}
\end{subtable}%
\hfill%
\begin{subtable}[t]{0.5\linewidth}
\centering
\begin{tabular}{l|cccc}
\toprule
Methods & Succ.$\uparrow$ & MPJPE$\downarrow$ & $E_\text{vel}\downarrow$ & $E_\text{acc}\downarrow$ \\
\midrule
OmniH2O~\citep{he2025omnih2o} & $72.2\%$ & $73.43$ & $11.78$ & $10.48$ \\
UH-1~\citep{11203143} & $68.8\%$ & $121.51$ & $16.59$ & $14.80$ \\
LangWBC$^*$~\citep{shao2025langwbc} & $76.0\%$ & $-$ & $-$ & $-$ \\
RLPF~\citep{yue2025rl} & $80.0\%$ & $140.00$ & $-$ & $-$ \\
UniAct-T2M~\citep{jiang2025uniact} & $82.5\%$ & $60.33$ & $9.15$ & $8.93$ \\
\midrule
\rowcolor[HTML]{EFEFEF}
Humanoid-LLA & $\mathbf{87.6\%}$ & $\mathbf{56.43}$ & $\mathbf{8.92}$ & $\mathbf{8.74}$ \\
\bottomrule
\end{tabular}
\end{subtable}

\vspace{-0.08in}
\label{tab:main_results}
\end{table}

\begin{table}[t]
\caption{Ablation results of CoT reasoning, RL finetuning, and physical reward terms.}
\centering
\resizebox{\linewidth}{!}{
\begin{tabular}{l|cccc||cccc}
\toprule
\multirow{1}{*}{Methods} &  \multirow{1}{*}{FID$\downarrow$}& \multirow{1}{*}{R-Precision$\uparrow$}         & \multirow{1}{*}{MM-Dist$\downarrow$} & \multirow{1}{*}{Div.$\rightarrow$}  & \multirow{1}{*}{Succ.$\uparrow$}  & \multirow{1}{*}{MPJPE$\downarrow$}  & \multirow{1}{*}{$\text{E}_\text{vel}\downarrow$}   & 
\multirow{1}{*}{$\text{E}_\text{acc}\downarrow$}          \\ \midrule

Humanoid-LLA w/o CoT            & $10.423$ & $0.270$ & $6.222$ & $6.405$ & $64.90\%$        &  $90.43$         & $14.11$     & $11.23$        \\
Humanoid-LLA w/o RLFT &            $5.132$        &     $0.331$        &     $5.443$   &    $6.668$            &     $68.64\%$               &      $78.31$           &      $12.12$  &    $10.01$      \\ 
Humanoid-LLA w/o $r_{\text{dist}}$              & $4.597$ & $0.342$ & $5.401$ & $6.892$  & $85.33\%$         & $61.27$         & $9.31$    &  $9.02$     \\

Humanoid-LLA w/o $r_{\text{track}}$      & $\mathbf{2.578}$ & ${0.439}$ & ${5.013}$ & ${7.007}$  & ${76.72\%}$      & ${66.42}$         & $10.89$      &  $9.77$   \\
\midrule
\rowcolor[HTML]{EFEFEF}
Humanoid-LLA (Ours)       & $2.626$ & $\mathbf{0.447}$ & $\mathbf{4.911}$ & $\mathbf{7.122}$  & $\mathbf{87.6\%}$         & $\mathbf{56.43}$         & $\mathbf{8.92}$   &   $\mathbf{8.74}$ \\
\bottomrule
\end{tabular}
}
\label{tab:ablation}
\vspace{-0.1in}
\end{table}

\subsection{Text-to-Humanoid Evaluation}
\label{exp: eval}

The results in Tab.~\ref{tab:main_results} reveal distinct trade-offs among baselines.  
MDM \citep{tevethuman} generates motions in the human domain and transfers them to robots via kinematic retargeting, preserving expressiveness and diversity but lacking physical fidelity.  
OmniH2O \citep{he2025omnih2o} adds an imitation policy to obtain feasible trajectories, yet discrepancies between human and robot action spaces cause frequent tracking failures, and discarding these biases the motion distribution.  
UH-1 \citep{11203143} trains on robot trajectories to decode from a robot-space latent manifold, improving fidelity and tracking scores while retaining generative capacity, but still falling short for real-world deployment.  
LangWBC \citep{shao2025langwbc} conditions on both language and control, achieving strong low-level executability but weaker motion--language alignment.  
RLPF \citep{yue2025rl} introduces physical feedback to constrain motions to the feasible set, but optimizing distributions in the human space yields suboptimal humanoid alignment.  
In contrast, our method couples LLM-generated tokens with a vocabulary-directed controller and fine-tunes with humanoid feedback, preserving diversity and expressiveness while substantially boosting physical fidelity. This leads to consistent improvements across both evaluation axes, outperforming prior methods on generation metrics and tracking metrics. 
Implementation details are in the Appendix.

\subsection{Ablation Studies}
We perform ablation studies to assess the contribution of each component of LLA in  terms of generation quality and physical fidelity. 
(1) \textbf{Humanoid-LLA w/o CoT}: removes chain-of-thought augmentation and relies solely on raw motion descriptions compared with the SFT baseline. 
(2) \textbf{Humanoid-LLA w/o RLFT}: replaces the RL fine-tuned model with the SFT-only baseline. 
(3) \textbf{Humanoid-LLA w/o $r_{\text{dist}}$}: excludes the distributional reward while retaining the tracking-based term. 
(4) \textbf{Humanoid-LLA w/o $r_{\text{track}}$}: excludes the tracking reward while retaining the distributional term. 
The results in Tab.~\ref{tab:ablation} highlight that the motion CoT is crucial for semantic planning: removing CoT sharply degrades text–motion alignment (FID 2.626 → 10.423; R-Precision 0.447 → 0.270) and also harms execution, suggesting that without explicit intermediate reasoning, the model struggles to translate some complex instructions into consistent token sequences. RLFT provides the missing physical grounding: compared to SFT-only, it substantially improves both executability and overall generation quality (Succ. 68.64\% → 87.6\%; MPJPE 78.31 → 56.43; FID 5.132 → 2.626), indicating that physics feedback corrects distributional mismatch that supervised learning alone cannot resolve. Reward ablations further reveal a clean separation: $r_{\text{dist}}$ mainly preserves semantic distribution (better FID/R-Precision), while $r_{\text{track}}$ is essential for stable execution (higher Succ. and lower MPJPE); combining them yields the best trade-off.

\section{Conclusion}
In this work, we present Humanoid-LLA, a unified framework for free-form language humanoid control that bridges high-level language commands and whole-body execution. Our approach addresses the critical challenges of language generalization and physical fidelity in text-to-humanoid whole body control. Specifically, Humanoid-LLA introduces a unified motion vocabulary that aligns human and humanoid motion primitives, effectively bridging large language models and a whole-body controller. By augmenting large-scale human-motion datasets with vision-language model-generated annotations and fine-tuning with humanoid feedback in simulation, our model achieves enhanced language generalization and physical feasibility at execution. Extensive evaluations on both physical feasibility and motion quality demonstrate that our method outperforms prior works in physical environments, culminating in successful deployment on real humanoid hardware. Extending Humanoid-LLA to richer multimodal grounding and longer-horizon planning remains an important direction.


\section*{Acknowledgement}
This work was supported by NSFC (No.62406195, W2431046), Shanghai Local College Capacity Building Program (23010503100), Shanghai Frontiers Science Center of Human-centered Artificial Intelligence (ShangHAI), MoE Key Laboratory of Intelligent Perception and Human-Machine Collaboration (ShanghaiTech University), the Shanghai Frontiers Science Center of Human-centered Artificial Intelligence, HPC Platform and Core Facility Platform of Computer Science and Communication of ShanghaiTech University and Shanghai Engineering Research Center of Intelligent Vision and Imaging.

\renewcommand{\bibfont}{\small}
\bibliographystyle{plainnat}
\bibliography{references}

@inproceedings{he2024learning,
  title={Learning human-to-humanoid real-time whole-body teleoperation},
  author={He, Tairan and Luo, Zhengyi and Xiao, Wenli and Zhang, Chong and Kitani, Kris and Liu, Changliu and Shi, Guanya},
  booktitle={2024 IEEE/RSJ International Conference on Intelligent Robots and Systems (IROS)},
  pages={8944--8951},
  year={2024},
  organization={IEEE}
}

@inproceedings{he2025omnih2o,
  title={OmniH2O: Universal and Dexterous Human-to-Humanoid Whole-Body Teleoperation and Learning},
  author={He, Tairan and Luo, Zhengyi and He, Xialin and Xiao, Wenli and Zhang, Chong and Zhang, Weinan and Kitani, Kris M and Liu, Changliu and Shi, Guanya},
  booktitle={Conference on Robot Learning},
  pages={1516--1540},
  year={2025},
  organization={PMLR}
}

@article{he2025asap,
  title={Asap: Aligning simulation and real-world physics for learning agile humanoid whole-body skills},
  author={He, Tairan and Gao, Jiawei and Xiao, Wenli and Zhang, Yuanhang and Wang, Zi and Wang, Jiashun and Luo, Zhengyi and He, Guanqi and Sobanbab, Nikhil and Pan, Chaoyi and others},
  journal={arXiv preprint arXiv:2502.01143},
  year={2025}
}

@INPROCEEDINGS{11203143,
      author={Mao, Jiageng and Zhao, Siheng and Song, Siqi and Hong, Chuye and Shi, Tianheng and Ye, Junjie and Zhang, Mingtong and Geng, Haoran and Malik, Jitendra and Guizilini, Vitor and Wang, Yue},
      booktitle={2025 IEEE-RAS 24th International Conference on Humanoid Robots (Humanoids)}, 
      title={Universal Humanoid Robot Pose Learning from Internet Human Videos}, 
      year={2025},
      volume={},
      number={},
      pages={1-8},
      doi={10.1109/Humanoids65713.2025.11203143}}

@article{shao2025langwbc,
  title={LangWBC: Language-directed Humanoid Whole-Body Control via End-to-end Learning},
  author={Shao, Yiyang and Huang, Xiaoyu and Zhang, Bike and Liao, Qiayuan and Gao, Yuman and Chi, Yufeng and Li, Zhongyu and Shao, Sophia and Sreenath, Koushil},
  journal={arXiv preprint arXiv:2504.21738},
  year={2025}
}

@inproceedings{shi2025almi,
  title={Adversarial Locomotion and Motion Imitation for Humanoid Policy Learning}, 
  author={Jiyuan Shi and Xinzhe Liu and Dewei Wang and Ouyang Lu and Sören Schwertfeger and Fuchun Sun and Chenjia Bai and Xuelong Li},
  booktitle={Neural Information Processing Systems (NeurIPS)},
  year={2025},
  url={https://arxiv.org/abs/2504.14305}, 
}

@article{yue2025rl,
  title={RL from Physical Feedback: Aligning Large Motion Models with Humanoid Control},
  author={Yue, Junpeng and Wang, Zepeng and Wang, Yuxuan and Zeng, Weishuai and Wang, Jiangxing and Xu, Xinrun and Zhang, Yu and Zheng, Sipeng and Ding, Ziluo and Lu, Zongqing},
  journal={arXiv preprint arXiv:2506.12769},
  year={2025}
}

@inproceedings{tevethuman,
  title={Human Motion Diffusion Model},
  author={Tevet, Guy and Raab, Sigal and Gordon, Brian and Shafir, Yoni and Cohen-or, Daniel and Bermano, Amit Haim},
  booktitle={The Eleventh International Conference on Learning Representations},
  year={2023}
}

@article{ze2025twist,
title={TWIST: Teleoperated Whole-Body Imitation System},
author= {Yanjie Ze and Zixuan Chen and João Pedro Araújo and Zi-ang Cao and Xue Bin Peng and Jiajun Wu and C. Karen Liu},
year= {2025},
journal= {arXiv preprint arXiv:2505.02833}
}

@article{ze2025twist2,
  title={TWIST2: Scalable, Portable, and Holistic Humanoid Data Collection System},
  author={Ze, Yanjie and Zhao, Siheng and Wang, Weizhuo and Kanazawa, Angjoo and Duan, Rocky and Abbeel, Pieter and Shi, Guanya and Wu, Jiajun and Liu, C Karen},
  journal={arXiv preprint arXiv:2511.02832},
  year={2025}
}

@article{ben2025homie,
  title={Homie: Humanoid loco-manipulation with isomorphic exoskeleton cockpit},
  author={Ben, Qingwei and Jia, Feiyu and Zeng, Jia and Dong, Junting and Lin, Dahua and Pang, Jiangmiao},
  journal={arXiv preprint arXiv:2502.13013},
  year={2025}
}

@inproceedings{zhang2023generating,
  title={T2M-GPT: Generating Human Motion from Textual Descriptions with Discrete Representations},
  author={Zhang, Jianrong and Zhang, Yangsong and Cun, Xiaodong and Huang, Shaoli and Zhang, Yong and Zhao, Hongwei and Lu, Hongtao and Shen, Xi},
  booktitle={Proceedings of the IEEE/CVF Conference on Computer Vision and Pattern Recognition (CVPR)},
  year={2023},
}

@article{ouyang2025motion,
  title={Motion-R1: Chain-of-Thought Reasoning and Reinforcement Learning for Human Motion Generation},
  author={Ouyang, Runqi and Li, Haoyun and Zhang, Zhenyuan and Wang, Xiaofeng and Zhu, Zheng and Huang, Guan and Wang, Xingang},
  journal={arXiv preprint arXiv:2506.10353},
  year={2025}
}

@article{yao2022controlvae,
  title={Controlvae: Model-based learning of generative controllers for physics-based characters},
  author={Yao, Heyuan and Song, Zhenhua and Chen, Baoquan and Liu, Libin},
  journal={ACM Transactions on Graphics (TOG)},
  volume={41},
  number={6},
  pages={1--16},
  year={2022},
  publisher={ACM New York, NY, USA}
}

@inproceedings{ross2011reduction,
  title={A reduction of imitation learning and structured prediction to no-regret online learning},
  author={Ross, St{\'e}phane and Gordon, Geoffrey and Bagnell, Drew},
  booktitle={Proceedings of the fourteenth international conference on artificial intelligence and statistics},
  pages={627--635},
  year={2011},
  organization={JMLR Workshop and Conference Proceedings}
}

@article{kingma2013auto,
  title={Auto-encoding variational bayes},
  author={Kingma, Diederik P and Welling, Max},
  journal={arXiv preprint arXiv:1312.6114},
  year={2013}
}

@article{van2017neural,
  title={Neural discrete representation learning},
  author={Van Den Oord, Aaron and Vinyals, Oriol and others},
  journal={Advances in neural information processing systems},
  volume={30},
  year={2017}
}

@article{unitok,
  title={UniTok: A Unified Tokenizer for Visual Generation and Understanding},
  author={Ma, Chuofan and Jiang, Yi and Wu, Junfeng and Yang, Jihan and Yu, Xin and Yuan, Zehuan and Peng, Bingyue and Qi, Xiaojuan},
  journal={arXiv preprint arXiv:2502.20321},
  year={2025}
}

@inproceedings{guo2022generating,
  title={Generating diverse and natural 3d human motions from text},
  author={Guo, Chuan and Zou, Shihao and Zuo, Xinxin and Wang, Sen and Ji, Wei and Li, Xingyu and Cheng, Li},
  booktitle={Proceedings of the IEEE/CVF conference on computer vision and pattern recognition},
  pages={5152--5161},
  year={2022}
}

@inproceedings{cui2024anyskill,
  title={Anyskill: Learning Open-Vocabulary Physical Skill for Interactive Agents},
  author={Cui, Jieming and Liu, Tengyu and Liu, Nian and Yang, Yaodong and Zhu, Yixin and Huang, Siyuan},
  booktitle={Conference on Computer Vision and Pattern Recognition(CVPR)},
  year={2024}
}

@inproceedings{punnakkal2021babel,
  title={BABEL: Bodies, action and behavior with english labels},
  author={Punnakkal, Abhinanda R and Chandrasekaran, Arjun and Athanasiou, Nikos and Quiros-Ramirez, Alejandra and Black, Michael J},
  booktitle={Proceedings of the IEEE/CVF conference on computer vision and pattern recognition},
  pages={722--731},
  year={2021}
}

@inproceedings{xu2024lagoon,
  title={LAGOON: Language-Guided Motion Control},
  author={Xu, Shusheng and Wang, Huaijie and Ouyang, Yutao and Gao, Jiaxuan and Mei, Zhiyu and Yu, Chao and Wu, Yi},
  booktitle={2024 IEEE International Conference on Robotics and Automation (ICRA)},
  pages={9743--9750},
  year={2024},
  organization={IEEE}
}

@article{jiang2024motiongpt,
  title={Motiongpt: Human motion as a foreign language},
  author={Jiang, Biao and Chen, Xin and Liu, Wen and Yu, Jingyi and Yu, Gang and Chen, Tao},
  journal={Advances in Neural Information Processing Systems},
  volume={36},
  year={2024}
}

@inproceedings{Luo2023PerpetualHC,
    author={Zhengyi Luo and Jinkun Cao and Alexander W. Winkler and Kris Kitani and Weipeng Xu},
    title={Perpetual Humanoid Control for Real-time Simulated Avatars},
    booktitle={International Conference on Computer Vision (ICCV)},
    year={2023}
}

@article{shao2024deepseekmath,
  title={Deepseekmath: Pushing the limits of mathematical reasoning in open language models},
  author={Shao, Zhihong and Wang, Peiyi and Zhu, Qihao and Xu, Runxin and Song, Junxiao and Bi, Xiao and Zhang, Haowei and Zhang, Mingchuan and Li, YK and others},
  journal={arXiv preprint arXiv:2402.03300},
  year={2024}
}

@article{xu2024humanvla,
  title={Humanvla: Towards vision-language directed object rearrangement by physical humanoid},
  author={Xu, Xinyu and Zhang, Yizheng and Li, Yong-Lu and Han, Lei and Lu, Cewu},
  journal={Advances in Neural Information Processing Systems},
  volume={37},
  pages={18633--18659},
  year={2024}
}

@inproceedings{kimopenvla,
  title={OpenVLA: An Open-Source Vision-Language-Action Model},
  author={Kim, Moo Jin and Pertsch, Karl and Karamcheti, Siddharth and Xiao, Ted and Balakrishna, Ashwin and Nair, Suraj and Rafailov, Rafael and Foster, Ethan P and Sanketi, Pannag R and Vuong, Quan and others},
  booktitle={8th Annual Conference on Robot Learning},
  year={2024}
}

@article{bjorck2025gr00t,
  title={Gr00t n1: An open foundation model for generalist humanoid robots},
  author={Bjorck, Johan and Casta{\~n}eda, Fernando and Cherniadev, Nikita and Da, Xingye and Ding, Runyu and Fan, Linxi and Fang, Yu and Fox, Dieter and Hu, Fengyuan and Huang, Spencer and others},
  journal={arXiv preprint arXiv:2503.14734},
  year={2025}
}

@article{Qwen2.5-VL,
  title={Qwen2.5-VL Technical Report},
  author={Bai, Shuai and Chen, Keqin and Liu, Xuejing and Wang, Jialin and Ge, Wenbin and Song, Sibo and Dang, Kai and Wang, Peng and Wang, Shijie and Tang, Jun and Zhong, Humen and Zhu, Yuanzhi and Yang, Mingkun and Li, Zhaohai and Wan, Jianqiang and Wang, Pengfei and Ding, Wei and Fu, Zheren and Xu, Yiheng and Ye, Jiabo and Zhang, Xi and Xie, Tianbao and Cheng, Zesen and Zhang, Hang and Yang, Zhibo and Xu, Haiyang and Lin, Junyang},
  journal={arXiv preprint arXiv:2502.13923},
  year={2025}
}

@inproceedings{chen2023executing,
  title={Executing your commands via motion diffusion in latent space},
  author={Chen, Xin and Jiang, Biao and Liu, Wen and Huang, Zilong and Fu, Bin and Chen, Tao and Yu, Gang},
  booktitle={Proceedings of the IEEE/CVF conference on computer vision and pattern recognition},
  pages={18000--18010},
  year={2023}
}

@article{zhang2024motiondiffuse,
  title={Motiondiffuse: Text-driven human motion generation with diffusion model},
  author={Zhang, Mingyuan and Cai, Zhongang and Pan, Liang and Hong, Fangzhou and Guo, Xinying and Yang, Lei and Liu, Ziwei},
  journal={IEEE transactions on pattern analysis and machine intelligence},
  volume={46},
  number={6},
  pages={4115--4128},
  year={2024},
  publisher={IEEE}
}

@inproceedings{yuan2023physdiff,
  title={Physdiff: Physics-guided human motion diffusion model},
  author={Yuan, Ye and Song, Jiaming and Iqbal, Umar and Vahdat, Arash and Kautz, Jan},
  booktitle={Proceedings of the IEEE/CVF international conference on computer vision},
  pages={16010--16021},
  year={2023}
}

@inproceedings{serifi2024robot,
  title={Robot motion diffusion model: Motion generation for robotic characters},
  author={Serifi, Agon and Grandia, Ruben and Knoop, Espen and Gross, Markus and B{\"a}cher, Moritz},
  booktitle={SIGGRAPH asia 2024 conference papers},
  pages={1--9},
  year={2024}
}

@inproceedings{karunratanakul2023guided,
  title={Guided motion diffusion for controllable human motion synthesis},
  author={Karunratanakul, Korrawe and Preechakul, Konpat and Suwajanakorn, Supasorn and Tang, Siyu},
  booktitle={Proceedings of the IEEE/CVF International Conference on Computer Vision},
  pages={2151--2162},
  year={2023}
}

@article{peng2018deepmimic,
  title={Deepmimic: Example-guided deep reinforcement learning of physics-based character skills},
  author={Peng, Xue Bin and Abbeel, Pieter and Levine, Sergey and Van de Panne, Michiel},
  journal={ACM Transactions On Graphics (TOG)},
  volume={37},
  number={4},
  pages={1--14},
  year={2018},
  publisher={ACM New York, NY, USA}
}

@article{peng2021amp,
  title={Amp: Adversarial motion priors for stylized physics-based character control},
  author={Peng, Xue Bin and Ma, Ze and Abbeel, Pieter and Levine, Sergey and Kanazawa, Angjoo},
  journal={ACM Transactions on Graphics (ToG)},
  volume={40},
  number={4},
  pages={1--20},
  year={2021},
  publisher={ACM New York, NY, USA}
}

@article{peng2022ase,
  title={Ase: Large-scale reusable adversarial skill embeddings for physically simulated characters},
  author={Peng, Xue Bin and Guo, Yunrong and Halper, Lina and Levine, Sergey and Fidler, Sanja},
  journal={ACM Transactions On Graphics (TOG)},
  volume={41},
  number={4},
  pages={1--17},
  year={2022},
  publisher={ACM New York, NY, USA}
}

@article{tessler2024maskedmimic,
  title={Maskedmimic: Unified physics-based character control through masked motion inpainting},
  author={Tessler, Chen and Guo, Yunrong and Nabati, Ofir and Chechik, Gal and Peng, Xue Bin},
  journal={ACM Transactions on Graphics (TOG)},
  volume={43},
  number={6},
  pages={1--21},
  year={2024},
  publisher={ACM New York, NY, USA}
}

@article{tessler2025maskedmanipulator,
  title={MaskedManipulator: Versatile Whole-Body Control for Loco-Manipulation},
  author={Tessler, Chen and Jiang, Yifeng and Coumans, Erwin and Luo, Zhengyi and Chechik, Gal and Peng, Xue Bin},
  journal={arXiv preprint arXiv:2505.19086},
  year={2025}
}

@article{wu2025uniphys,
  title={UniPhys: Unified Planner and Controller with Diffusion for Flexible Physics-Based Character Control},
  author={Wu, Yan and Karunratanakul, Korrawe and Luo, Zhengyi and Tang, Siyu},
  journal={arXiv preprint arXiv:2504.12540},
  year={2025}
}

@inproceedings{tevetclosd,
  title={CLoSD: Closing the Loop between Simulation and Diffusion for multi-task character control},
  author={Tevet, Guy and Raab, Sigal and Cohan, Setareh and Reda, Daniele and Luo, Zhengyi and Peng, Xue Bin and Bermano, Amit Haim and van de Panne, Michiel},
  booktitle={The Thirteenth International Conference on Learning Representations},
  year={2024}
}

@article{yao2024moconvq,
  title={Moconvq: Unified physics-based motion control via scalable discrete representations},
  author={Yao, Heyuan and Song, Zhenhua and Zhou, Yuyang and Ao, Tenglong and Chen, Baoquan and Liu, Libin},
  journal={ACM Transactions on Graphics (TOG)},
  volume={43},
  number={4},
  pages={1--21},
  year={2024},
  publisher={ACM New York, NY, USA}
}

@inproceedings{he2025hover,
  title={Hover: Versatile neural whole-body controller for humanoid robots},
  author={He, Tairan and Xiao, Wenli and Lin, Toru and Luo, Zhengyi and Xu, Zhenjia and Jiang, Zhenyu and Kautz, Jan and Liu, Changliu and Shi, Guanya and Wang, Xiaolong and others},
  booktitle={2025 IEEE International Conference on Robotics and Automation (ICRA)},
  pages={9989--9996},
  year={2025},
  organization={IEEE}
}

@article{cheng2024expressive,
  title={Expressive whole-body control for humanoid robots},
  author={Cheng, Xuxin and Ji, Yandong and Chen, Junming and Yang, Ruihan and Yang, Ge and Wang, Xiaolong},
  journal={arXiv preprint arXiv:2402.16796},
  year={2024}
}

@article{ji2024exbody2,
  title={ExBody2: Advanced Expressive Humanoid Whole-Body Control}, 
  author={Ji, Mazeyu and Peng, Xuanbin and Liu, Fangchen and Li, Jialong and Yang, Ge and Cheng, Xuxin and Wang, Xiaolong},
  journal={arXiv preprint arXiv:2412.13196},
  year={2024},
  }

@article{li2025amo,
title={AMO: Adaptive Motion Optimization for Hyper-Dexterous Humanoid Whole-Body Control},
author={Li, Jialong and Cheng, Xuxin and Huang, Tianshu and Yang, Shiqi and Qiu, Rizhao and Wang, Xiaolong},
journal={Robotics: Science and Systems 2025},
year={2025}
}

@misc{li2025clone,
          title={CLONE: Closed-Loop Whole-Body Humanoid Teleoperation for Long-Horizon Tasks}, 
          author={Yixuan Li and Yutang Lin and Jieming Cui and Tengyu Liu and Wei Liang and Yixin Zhu and Siyuan Huang},
          journal={arXiv preprint arXiv:2506.08931}, 
          year={2025}
        }

@article{yin2025unitracker,
  title={UniTracker: Learning Universal Whole-Body Motion Tracker for Humanoid Robots},
  author={Yin, Kangning and Zeng, Weishuai and Fan, Ke and Wang, Zirui and Zhang, Qiang and Tian, Zheng and Wang, Jingbo and Pang, Jiangmiao and Zhang, Weinan},
  journal={arXiv preprint arXiv:2507.07356},
  year={2025}
}

@article{joao2025gmr,
title={Retargeting Matters: General Motion Retargeting for Humanoid Motion Tracking},
author= {Joao Pedro Araujo and Yanjie Ze and Pei Xu and Jiajun Wu and C. Karen Liu},
year= {2025},
journal= {arXiv preprint arXiv:2510.02252}
}

@article{liao2025beyondmimic,
  title={Beyondmimic: From motion tracking to versatile humanoid control via guided diffusion},
  author={Liao, Qiayuan and Truong, Takara E and Huang, Xiaoyu and Tevet, Guy and Sreenath, Koushil and Liu, C Karen},
  journal={arXiv e-prints},
  pages={arXiv--2508},
  year={2025}
}

@inproceedings{juravsky2022padl,
  title={Padl: Language-directed physics-based character control},
  author={Juravsky, Jordan and Guo, Yunrong and Fidler, Sanja and Peng, Xue Bin},
  booktitle={SIGGRAPH Asia 2022 Conference Papers},
  pages={1--9},
  year={2022}
}

@inproceedings{truong2024pdp,
  title={Pdp: Physics-based character animation via diffusion policy},
  author={Truong, Takara Everest and Piseno, Michael and Xie, Zhaoming and Liu, Karen},
  booktitle={SIGGRAPH Asia 2024 Conference Papers},
  pages={1--10},
  year={2024}
}

@inproceedings{juravsky2024superpadl,
  title={Superpadl: Scaling language-directed physics-based control with progressive supervised distillation},
  author={Juravsky, Jordan and Guo, Yunrong and Fidler, Sanja and Peng, Xue Bin},
  booktitle={ACM SIGGRAPH 2024 Conference Papers},
  pages={1--11},
  year={2024}
}

@article{zhao2025smap,
  title={SMAP: Self-supervised Motion Adaptation for Physically Plausible Humanoid Whole-body Control},
  author={Zhao, Haoyu and Lin, Sixu and Ben, Qingwei and Dai, Minyue and Fei, Hao and Wang, Jingbo and Zou, Hua and Dong, Junting},
  journal={arXiv preprint arXiv:2505.19463},
  year={2025}
}

@conference{AMASS:ICCV:2019,
  title = {{AMASS}: Archive of Motion Capture as Surface Shapes},
  author = {Mahmood, Naureen and Ghorbani, Nima and Troje, Nikolaus F. and Pons-Moll, Gerard and Black, Michael J.},
  booktitle = {International Conference on Computer Vision},
  pages = {5442--5451},
  month = oct,
  year = {2019},
  month_numeric = {10}
}

@misc{zakka10mink,
  author       = {Kevin Zakka},
  title        = {Mink: Python inverse kinematics based on MuJoCo},
  year         = {2024},
  howpublished = {\url{https://github.com/kevinzakka/mink}}
}

@inproceedings{han2025reindiffuse,
  title={Reindiffuse: Crafting physically plausible motions with reinforced diffusion model},
  author={Han, Gaoge and Liang, Mingjiang and Tang, Jinglei and Cheng, Yongkang and Liu, Wei and Huang, Shaoli},
  booktitle={2025 IEEE/CVF Winter Conference on Applications of Computer Vision (WACV)},
  pages={2218--2227},
  year={2025},
  organization={IEEE}
}

@article{schulman2017proximal,
  title={Proximal policy optimization algorithms},
  author={Schulman, John and Wolski, Filip and Dhariwal, Prafulla and Radford, Alec and Klimov, Oleg},
  journal={arXiv preprint arXiv:1707.06347},
  year={2017}
}

@article{xue2025leverb,
  title={Leverb: Humanoid whole-body control with latent vision-language instruction},
  author={Xue, Haoru and Huang, Xiaoyu and Niu, Dantong and Liao, Qiayuan and Kragerud, Thomas and Gravdahl, Jan Tommy and Peng, Xue Bin and Shi, Guanya and Darrell, Trevor and Sreenath, Koushil and others},
  journal={arXiv preprint arXiv:2506.13751},
  year={2025}
}

@article{ding2025humanoid,
  title={Humanoid-vla: Towards universal humanoid control with visual integration},
  author={Ding, Pengxiang and Ma, Jianfei and Tong, Xinyang and Zou, Binghong and Luo, Xinxin and Fan, Yiguo and Wang, Ting and Lu, Hongchao and Mo, Panzhong and Liu, Jinxin and others},
  journal={arXiv preprint arXiv:2502.14795},
  year={2025}
}

@article{yin2025visualmimic,
title={VisualMimic: Visual Humanoid Loco-Manipulation via Motion Tracking and Generation},
author= {Shaofeng Yin and Yanjie Ze and Hong-Xing Yu and C. Karen Liu and Jiajun Wu},
year= {2025},
journal= {arXiv preprint arXiv:2509.20322}
}

@inproceedings{hwangsnapmogen,
  title={SnapMoGen: Human Motion Generation from Expressive Texts},
  author={Hwang, Inwoo and Wang, Jian and Zhou, Bing and others},
  booktitle={The Thirty-ninth Annual Conference on Neural Information Processing Systems},
  year={2025}
}

@article{lin2023motion,
  title={Motion-x: A large-scale 3d expressive whole-body human motion dataset},
  author={Lin, Jing and Zeng, Ailing and Lu, Shunlin and Cai, Yuanhao and Zhang, Ruimao and Wang, Haoqian and Zhang, Lei},
  journal={Advances in Neural Information Processing Systems},
  volume={36},
  pages={25268--25280},
  year={2023}
}

@inproceedings{fan2025go,
  title={Go to zero: Towards zero-shot motion generation with million-scale data},
  author={Fan, Ke and Lu, Shunlin and Dai, Minyue and Yu, Runyi and Xiao, Lixing and Dou, Zhiyang and Dong, Junting and Ma, Lizhuang and Wang, Jingbo},
  booktitle={Proceedings of the IEEE/CVF International Conference on Computer Vision},
  pages={13336--13348},
  year={2025}
}

@article{jiang2025uniact,
  title={UniAct: Unified Motion Generation and Action Streaming for Humanoid Robots},
  author={Jiang, Nan and He, Zimo and Yu, Wanhe and Pang, Lexi and Li, Yunhao and Li, Hongjie and Cui, Jieming and Li, Yuhan and Wang, Yizhou and Zhu, Yixin and others},
  journal={arXiv preprint arXiv:2512.24321},
  year={2025}
}

@article{li2025language,
  title={From language to locomotion: Retargeting-free humanoid control via motion latent guidance},
  author={Li, Zhe and Chi, Cheng and Wei, Yangyang and Zhu, Boan and Peng, Yibo and Huang, Tao and Wang, Pengwei and Wang, Zhongyuan and Zhang, Shanghang and Xu, Chang},
  journal={arXiv preprint arXiv:2510.14952},
  year={2025}
}

@article{li2026w1,
  title={FRoM-W1: Towards General Humanoid Whole-Body Control with Language Instructions},
  author={Li, Peng and Zhuang, Zihan and Gao, Yangfan and Dong, Yi and Li, Sixian and Jiang, Changhao and Dou, Shihan and Xi, Zhiheng and Zhou, Enyu and Huang, Jixuan and others},
  journal={arXiv preprint arXiv:2601.12799},
  year={2026}
}

@article{luo2025sonic,
  title={Sonic: Supersizing motion tracking for natural humanoid whole-body control},
  author={Luo, Zhengyi and Yuan, Ye and Wang, Tingwu and Li, Chenran and Chen, Sirui and Casta{\~n}eda, Fernando and Cao, Zi-Ang and Li, Jiefeng and Minor, David and Ben, Qingwei and others},
  journal={arXiv preprint arXiv:2511.07820},
  year={2025}
}

@article{wang2025sentinel,
  title={Sentinel: A fully end-to-end language-action model for humanoid whole body control},
  author={Wang, Yuxuan and Jiang, Haobin and Yao, Shiqing and Ding, Ziluo and Lu, Zongqing},
  journal={arXiv preprint arXiv:2511.19236},
  year={2025}
}

@article{yang2025qwen3,
  title={Qwen3 technical report},
  author={Yang, An and Li, Anfeng and Yang, Baosong and Zhang, Beichen and Hui, Binyuan and Zheng, Bo and Yu, Bowen and Gao, Chang and Huang, Chengen and Lv, Chenxu and others},
  journal={arXiv preprint arXiv:2505.09388},
  year={2025}
}

@article{hymotion2025,
  title={HY-Motion 1.0: Scaling Flow Matching Models for Text-To-Motion Generation},
  author={Tencent Hunyuan 3D Digital Human Team},
  journal={arXiv preprint arXiv:2512.23464},
  year={2025}
}

@article{yang2025omniretarget,
  title={Omniretarget: Interaction-preserving data generation for humanoid whole-body loco-manipulation and scene interaction},
  author={Yang, Lujie and Huang, Xiaoyu and Wu, Zhen and Kanazawa, Angjoo and Abbeel, Pieter and Sferrazza, Carmelo and Liu, C Karen and Duan, Rocky and Shi, Guanya},
  journal={arXiv preprint arXiv:2509.26633},
  year={2025}
}

@inproceedings{jiang2025harmon,
  title={Harmon: Whole-Body Motion Generation of Humanoid Robots from Language Descriptions},
  author={Jiang, Zhenyu and Xie, Yuqi and Li, Jinhan and Yuan, Ye and Zhu, Yifeng and Zhu, Yuke},
  booktitle={Conference on Robot Learning},
  pages={3015--3026},
  year={2025},
  organization={PMLR}
}

@article{xie2026textop,
  title={Textop: Real-time interactive text-driven humanoid robot motion generation and control},
  author={Xie, Weiji and Zheng, Jiakun and Han, Jinrui and Shi, Jiyuan and Zhang, Weinan and Bai, Chenjia and Li, Xuelong},
  journal={arXiv preprint arXiv:2602.07439},
  year={2026}
}

@article{bao2026phygile,
  title={PhyGile: Physics-Prefix Guided Motion Generation for Agile General Humanoid Motion Tracking},
  author={Bao, Jiacheng and Yang, Haoran and Xin, Yucheng and Liu, Junhong and Xu, Yuecheng and Liang, Han and Han, Pengfei and Ma, Xiaoguang and Wang, Dong and Zhao, Bin},
  journal={arXiv preprint arXiv:2603.19305},
  year={2026}
}

@article{yuan2026roboforge,
  title={RoboForge: Physically Optimized Text-guided Whole-Body Locomotion for Humanoids},
  author={Yuan, Xichen and Li, Zhe and Lyu, Bofan and Zuo, Kuangji and Lu, Yanshuo and Li, Gen and Yang, Jianfei},
  journal={arXiv preprint arXiv:2603.17927},
  year={2026}
}
\clearpage






\end{document}